%% file: abstract.tex
\documentclass[sigconf]{acmart}

\def\BibTeX{{\rm B\kern-.05em{\sc i\kern-.025em b}\kern-.08em
    T\kern-.1667em\lower.7ex\hbox{E}\kern-.125emX}}

\usepackage{booktabs} 

\usepackage{array}
\usepackage{float}
\usepackage{epsfig}
\usepackage{makecell}
\usepackage{adjustbox}

\usepackage{hyperref}
\usepackage{caption}
\usepackage{subcaption}
\usepackage{cleveref}
\usepackage{lipsum,graphicx,multicol}
\graphicspath{ {./images/} }
 
\setcopyright{acmcopyright}
\copyrightyear{2021}
\acmYear{2021}
\acmDOI{10.1145/1122445.1122456}

\acmConference[AutoML 2021]{The Fifth International Workshop on Automation in Machine Learning(KDD 2021)}{August 14-18, 2021}{Virtual, Singapore} 

\acmISBN{978-1-4503-XXXX-X/18/06}
\begin{document}
\title{To tune or not to tune? An Approach for Recommending Important Hyperparameters}

\author{Mohamadjavad Bahmani}
\orcid{1234-5678-9012}
\affiliation{%
  \institution{Data Systems Group, University of Tartu}
  \city{Tartu} 
  \country{Estonia} 
  \postcode{51009}
}
\email{{firstname.lastname}@ut.ee}

\author{Radwa El Shawi}
 
\affiliation{%
  \institution{Data Systems Group, University of Tartu}
  \city{Tartu} 
  \country{Estonia} 
  \postcode{51009}
}
\email{{firstname.lastname}@ut.ee}

\author{Nshan Potikyan}
\affiliation{%
  \institution{Data Systems Group, University of Tartu}
  \city{Tartu} 
  \country{Estonia}}
  \postcode{51009}
\email{{firstname.lastname}@ut.ee}

\author{Sherif Sakr}
\affiliation{%
  \institution{Data Systems Group, University of Tartu}
  \city{Tartu} 
  \country{Estonia}}
  \postcode{51009}
\email{{firstname.lastname}@ut.ee}

\renewcommand{\shortauthors}{}

\begin{abstract}
Novel technologies in automated machine learning ease the complexity of algorithm selection and hyperparameter optimization.
Hyperparameters are important for machine learning models as they significantly influence the  performance of machine learning models. Many optimization techniques have achieved notable success in hyperparameter tuning and surpassed the performance of human experts. However, depending on such techniques as black-box algorithms can leave machine learning practitioners without insight into the relative importance of different hyperparameters.  In this paper, we consider building the relationship between the performance of the machine learning models and their hyperparameters to discover the trend and gain insights, with empirical results based on six classifiers and 200 datasets. Our results enable users to decide whether it is worth conducting a possibly time-consuming tuning strategy, to focus on the most important hyperparameters, and to choose adequate hyperparameter spaces for tuning. The results of our experiments show that gradient boosting and Adaboost outperform other classifiers across 200 problems. However, they need tuning to boost their performance. Overall, the results obtained from this study provide a quantitative basis to focus efforts toward guided automated hyperparameter optimization and contribute toward the development of better-automated machine learning frameworks. 
\end{abstract}

%
%




\maketitle

\input{samplebody-conf}


\bibliographystyle{ACM-Reference-Format}
\bibliography{sample-base}

%
% ...

\end{document}

%% file: samplebody-conf.tex
\section{Introduction}
Machine learning has achieved notable success in different application domains including, image classification, object detection, self-driving cars, financial applications, recommendation systems, and medical diagnosis systems~\cite{krizhevsky2017imagenet,he2016deep,redmon2016you,rajkomar2019machine}. There is no doubt that we are living in the era of big data and, witnessing huge expansion in the amount of data generated every day from different resources. Such data should be analysed for richer and more robust insights. As the number of data scientists cannot scale with the number of applications, a novel research area has emerged, commonly referred to as Automated Machine Learning (AutoML). AutoML aims to automatically determine the approach that performs best for this particular application~\cite{hutter2019automated}. Thereby, AutoML makes state-of-the-art machine learning approaches accessible to non-experts who are interested in applying machine learning but do not have the knowledge and resources needed to build machine learning pipelines.

Every machine learning model has a set of hyperparameters, and the most basic task in AutoML is to tune these hyperparameters to achieve the best performance automatically. Hence, tuning hyperparameters becomes a key problem for a machine learning model. Hyperparameter tuning is an optimization problem where the objective function of optimization is a black-box function. Traditional optimization techniques such as grid search suffer from scalability problems and thus, there has been a recent surge of interest in more efficient optimization techniques~\cite{hutter2011sequential,bergstra2012random,bergstra2011algorithms}. The considerable success achieved by such techniques has not been accompanied by sufficient effort in providing answers to a set of questions such as how important each hyperparameter is to the model? What is the best value for each of these hyperparameters? Which hyperparameter interactions matter? How tunable is the machine learning algorithm? How the answers to these questions relate to the characteristics of the dataset under consideration? Answering these questions will improve the overall efficiency of hyperparameter optimization by reducing the search of hyperparameters that are important to the model. Therefore, performing automatic tuning while achieving high precision and high efficiency is an open problem that has not been fully addressed in machine learning. 

In this paper, we aim to fill this gap from a statistical point of view to simplify the tuning process for laying machine learning practitioners, and to optimize decision-making for more advanced processes. More specifically, we present a methodology to determine the hyperparameters that appear important to the model, in addition to robust defaults and well-performing machine learning algorithms based on empirical performance data derived from experiments across various datasets. We apply our approach on a benchmark study
of six popular classification algorithms: random forests~\cite{breiman2001random}, support vector machines(SVMs)~\cite{chang2011libsvm}, Adaboost~\cite{freund1997decision}, Extra Trees~\cite{geurts2006extremely}, decision tree~\cite{loh2011classification} and gradient boosting~\cite{friedman2001greedy}. More specifically, we analyze the importance of
hyperparameter tuning based on the performance of these algorithms across 200 datasets obtained from the OpenML ~\cite{vanschoren2014openml}. To ensure repeatability as one of the main targets of this work, we provide access to the source codes and the detailed results for the experiments of our study.\footnote{https://github.com/DataSystemsGroupUT/HyperParameterTunability} 

The remainder of this work is organized as follows. In Section~\ref{Sec:Relatedwork}, we provide a short overview of related work. In Section~\ref{Sec:FunctionalANOVA}, we cover the background required about functional ANOVA used in our analysis. Section~\ref{Sec:SetUP} covers details of experiments, including the used datasets, algorithms, hyperparameters, performance measures, and search techniques. Section~\ref{Sec:Results} describes the details of our experiments before we conclude the paper in Section~\ref{SEC:Conclusion}.

\section{Related work}\label{Sec:Relatedwork}
To the best of our knowledge, a very limited number of research papers consider the problem of tunability of machine learning search spaces. Bergstra and Bengio~\cite{bergstra2012random} studied the importance of neural networks hyperparameters and concluded that some hyperparameters are important to the model across all datasets, while others are important on some datasets. Their findings were used as an argument to justify why random search outperforms grid search when setting hyperparameters for neural networks. Mantovani et al.~\cite{mantovani2016hyper} applied different techniques to tune the hyperparameters of the decision tree algorithm on 102 heterogeneous datasets. The experimental results show that the improvement of the tuned models over all datasets, in most cases, is statistically significant. 

Different approaches exist for evaluating the importance of hyperparameters. The principle of forward selection has been used to predict a classifier performance based on a subset hyperparameters that is initialized as empty and greedily updated with the most important hyperparameter~\cite{hutter2013identifying}. In the same vein, Fawcett and Hoos ~\cite{fawcett2016analysing} presented ablation analysis, a technique for identifying hyperparameters that mostly contribute to improving the performance after tuning. For each of the considered hyperparameters, the performance improvement is computed by changing its value from a source configuration to a specific destination configuration specified by the tuning strategy (e.g., a user-defined default configuration and one obtained using automated configuration). Hutter et al.~\cite{hutter2013identifying} introduced an approach for assessing the importance of a single hyperparameter based on functional ANOVA~\cite{sobol1993sensitivity}, which is a powerful framework that can detect the importance of both individual hyperparameters and interaction effects between arbitrary subsets of hyperparameters. More specifically, functional ANOVA decomposes the variance in model performance into adaptive components due to all subsets of the hyperparameters. Estimation of Distribution Algorithms (EDA)~\cite{larranaga2001estimation} used to estimate the best default value for hyperparameters over the performance data by fitting a probability distribution to points in the input space and using this probability distribution to sample new points from.

Another line of research is based on meta-learning and performance data from similar datasets, and the resulting predictions are used to recommend a particular set of configurations for the new dataset. These techniques have achieved great success in recommending good hyperparameters~\cite{miranda2014hybrid,soares2004meta} to warm-start different optimization techniques~\cite{feurer2015initializing} or prune search spaces~\cite{wistuba2015hyperparameter}. The main drawback of these techniques is that it is  difficult to choose appropriate meta-features. In addition, extracting such meta-features is a time-consuming process. Another meta-learning approach that alleviates meta-features is Multi-task Bayesian optimization~\cite{swersky2013multi}. A multi-task model is built on the outcome of classifiers to find correlations between tasks which can be used to recommend hyperparameters for a task. However, this technique suffers from the cubic time complexity.

In this study, we apply a technique for quantifying the importance of the hyperparameters of machine learning algorithms. Following Hutter et al.~\cite{van2017empirical}, we consider a setting slightly more general by considering a larger number of machine learning algorithms across a large number of datasets. To the best of our knowledge, all the techniques considered in the literature either applied to a limited number of machine learning algorithms and datasets or relied on defaults as a reference point. In order to obtain representative results, we analyze hyperparameter importance across many different datasets. More specifically, we employ the 200 datasets from the OpenML ~\cite{vanschoren2014openml} to determine the most important hyperparameters of six classifiers along with the best defaults.
 
 \section{Functional ANOVA for estimating hyperparameter importance}\label{Sec:FunctionalANOVA}
The functional ANOVA framework is an efficient technique for assessing the importance of hyperparameters of a machine learning algorithm based on the efficient computations of marginal performance. More specifically, Functional ANOVA specifies the contribution of each hyperparameter to the variance of the machine learning algorithm performance. In the following we give a quick overview on how functional ANOVA is used to efficiently compute the importance of all hyperparameters. For more details about the efficient computation of functional ANOVA, we refer the interested reader to the original work done by Hutter et al. ~\cite{hutter2014efficient}.

Given an algorithm $A$ with $n$ hyperparameters with domain $\Theta_{1}, \Theta_{2},..., \Theta{n}$ and configuration space $\Theta={\Theta_{1}\times \Theta_{2} ...\times \Theta{n}}$. A configuration of algorithm $A$ is a vector $\theta=\langle \theta_{1}, \theta_{2},...,\theta_{n} \rangle$, where $\theta_{i} \in{\Theta_{i}}$. Let a partial configuration of $A$ is defined as a vector $\theta_{U}=\langle \theta_{1}, \theta_{2},...\theta_{j} \rangle$ with $U$ fixed hyperparameters, where $U$ is a subset of the set of all hyperparameters $N$ of algorithm $A$. The \emph{marginal performance} 
$\hat{a}_{U}(\theta_{U})$ is defined to be the average performance of algorithm $A$ for all complete configurations $\theta$ that have in common $\theta_{U}$. Computing such $\hat{a}_{U}(\theta_{U})$ is computationally expensive, however it has been shown by Hutter el al.~\cite{hutter2014efficient} that for tree-based model, $\hat{a}_{U}(\theta_{U})$ can be computed by a produce that is linear in the number of leaves in the model.

We apply functional ANOVA on each of the six classifiers as follows. First we collect performance data $\langle \theta_{i},y_{i}\rangle^{K}_{k=1}$ for each algorithm $A$ with $K$ different configurations, where $y_{i}$ is the performance of algorithm $A$ measured by Area Under the Curve (AUC). Next, we fit a random forest model to the performance data and then use functional ANOVA to decompose the variance in performance of the random forest $\hat{y}: \Theta_{1}\times \Theta_{2} ...\Theta_{n} -> \mathbb{R}$ into additive components that depends on subsets of the hyperparameters $N$:

\begin{equation}
\hat{y}(\theta)= \sum_{U\subseteq N} \hat{f_{U}}(\theta_{U})
\end{equation}

where the components $\hat{f_{U}}(\theta_{U})$ are defined as follows:

\begin{equation}
  \hat{f_{U}}(\theta_{U}) =
    \begin{cases}
      \hat{f_{\emptyset}} & \text{if $U=\emptyset$} \\
      \hat{a}_{U}(\theta_{U})-\sum_{W\subsetneq U} \hat{f_{W}}(\theta_{W}) & \text{otherwise,}
    \end{cases}       
\end{equation}

where $\hat{f}_{\emptyset}$ is the mean value of the function $\hat{y}$ over its domain. The unary function $\hat{f_{\{j\}}}(\theta_{\{j\}})$ captures the importance of hyperparameter $j$ average over all possible values for the rest of the hyperparameters, while $\hat{f_{\{U\}}}(\theta_{\{U\}})$ captures the interaction effects between all hyperparameters in $U$. Functional ANOVA decomposes the variance $V$ in the $\hat{y}$ into the contributions of all possible subsets of hyperparameters $V_{U}$ of algorithm $A$.

\begin{equation}
V= \sum_ {U\subseteq N} V_{U}\text{, }  \text{ where }  V_{U}=\frac{1}{||\Theta_{U}||}\int \hat{f_{U}}(\theta_{U})^2 d\theta_{U} 
\end{equation}

The importance of a hyperparameter or a set of hyperparameters is captured by the fraction of the variance the hyperparameter or the set of hyperparameters is responsible for; the higher the fraction, the more important the hyperparameter or the set of hyperparameters is to the model. Thus, such a hyperparameter should be tuned in order to achieve a good performance.
 
\begin{table*}[!ht]
\caption{Hyperparameters of the six machine learning algorithms used in this study.}
\label{tab:hyperparameters_details}
\begin{adjustbox}{max width=\textwidth}
    \centering
    \begin{tabular}{|c|c|c|c|} 
    
    \hline
    Classifiers & Hyperparameters & Values & Description  \\ 
    \hline

    SVMs
    & \thead{ C \\ kernel \\ coef0 \\ gamma \\ imputation \\ shrinking \\ tol}
    & \thead{$[2^{-5},2^{15}] $ \\ $['sigmoid', 'rbf']$ \\ $[-1,1]$ \\ $[2^{-15},2^3] $ \\ {mean, median, mode} \\ {true, false} \\ $[10^{-5},10^{-1}] $ }  
    & \thead{ complexity or C is used for regularization. \\ The type of the kernel function(due to curse of implementation we used just two main Kernel types). \\ A coefficient which impacts just on the sigmoid kernel. \\ It depends on the type of kernel   \\ The methods of imputations if the dataset has the missing values. \\ Utilization of shrinking heuristic or not. ( refer to \cite{joachims1998making}) \\ it represents acceptable tolerance for stopping criterion}  \\   
    \hline

    Random Forest 
    & \thead{bootstrap \\ max. features \\ min. samples leaf \\ min. samples split \\ imputation \\ criterion}
    & \thead{ {true, false} \\ $[0.1, 0.9]$ \\ $[1, 20] $ \\ $[2, 20]$ \\ {mean, median, mode} \\ {entropy, gini} }  
    & \thead{If True, bootstrap samples are used for creating trees otherwise the whole dataset.  \\ The subset of features when trying to find the best split. \\ It specifies the minimum number of data points for making a leaf.  \\ It specifies the minimum number of data points for making a split (internal node)\\ The methods of imputation if the dataset has the missing values. \\ It defines how to measure the quality of a split.}  \\    
    \hline

    Adaboost 
    & \thead{algorithm \\ imputation \\ n. estimator \\ learning rate \\ max. depth }
    & \thead{ {SAMME, SAMME.R} \\ {mean, median, mode} \\ $[50, 500]$ \\ $[0.01, 2.0]$ (log-scale)\\ $[1, 10]$ \\  }  
    & \thead{Selecting the type of boosting algorithm.(refer to \cite{hastie2009multi}) \\ The methods of imputation if the dataset has the missing values. \\ The maximum number of weak learners. \\ It controls the contributions of the weak learners. \\ It defines the maximum depth for each weak learner.}  \\  
    \hline

    Extra Trees 
    & \thead{Bootstrap \\ imputation \\  Criterion \\ max. features \\ min. samples split \\ min. samples leaf  }
    & \thead{ {true, false} \\ {mean, median, mode} \\ {entropy, gini} \\ $[0.1, 0.9]$ \\ $[2, 20]$ \\ $[1, 20]$} 
    & \thead{If True, bootstrap samples are used for creating trees otherwise the whole dataset. \\ The methods of imputation if the dataset has the missing values. \\ It defines how to measure the quality of a split. \\ The subset of features when trying to find the best split. \\  It specifies the minimum number of data points for making a split (internal node) \\  It specifies the minimum number of data points for making a leaf.}  \\  
    \hline
    
     Decision Tree 
    & \thead{imputation \\  Criterion \\ max. features \\ min. samples split \\ min. samples leaf  }
    & \thead{ {mean, median, mode} \\ {entropy, gini} \\ $[0.1, 0.9]$ \\ $[2, 20]$ \\ $[1, 20]$} 
    & \thead{ The methods of imputation if the dataset has the missing values.\\  It defines how to measure the quality of a split. \\  The subset of features when trying to find the best split. \\  It specifies the minimum number of data points for making a split (internal node) \\ It specifies the minimum number of data points for making a leaf.}  \\  
    \hline

    Gradient Boosting 
    & \thead{imputation \\ Learning rate \\ Criterion \\ n. estimators \\ max. depth \\ min. sample split \\ min. samples leaf \\ max. features }
    
    & \thead{ {mean, median, mode} \\ $[0.01, 0.99]$ \\ {friedman\_mse, mse} \\ $[50, 500]$ \\ $[1, 10]$ \\ $[2, 20]$ \\ $[1, 20]$ \\ $[0.1, 0.9]$} 
     
    & \thead{ The methods of imputation if the dataset has the missing values. \\  It controls the contributions of the weak learners. \\ The function to measure the quality of a split. \\ The maximum number of weak learners. \\  It defines the maximum depth for each weak learner. \\  It specifies the minimum number of data points for making a split (internal node) \\  It specifies the minimum number of data points for making a leaf. \\  The subset of features when trying to find the best split.}  \\   
     \hline

\end{tabular}

\end{adjustbox}
\end{table*}
\section{Experimental setup} \label{Sec:SetUP}
In this study, we use 200 datasets from the OpenML project~\cite{vanschoren2014openml} from various domains. The datasets containing between 500 to 100,000 data points are well-balanced and have been used in relevant scientific publications. These criteria ensure that the classification tasks are challenging, meaningful, and the results obtained from such tasks are comparable to earlier studies. The datasets are preprocessed in the same way as follows. The missing values are imputed with the mode for categorical features. For numerical features, the imputation is done using mean, median and mode. For SVMs classifier, the input variables are scaled to have unit variance as SVMs is sensitive to the scale of the input variables. 

We analyze the hyperparameters of six classifiers implemented on scikit-learn~\cite{buitinck2013api,pedregosa2011scikit}, namely, random forests (RF), support vector machines (SVMs), Adaboost (AB), Extra Trees (ET), decision tree (DT), and gradient boosting (GB). We consider the same hyperparameters and ranges used by Auto-sklearn~\cite{feurer2019auto}. Table~\ref{tab:hyperparameters_details} shows the considered hyperparameters for each of the considered classifiers along with their description, ranges and potential transformation function. We build a knowledge base that contains performance data of 500 different configurations for each of the six used classifiers per dataset. Each record in the knowledge base represents a 10-fold cross-validated evaluation of a certain configuration on a dataset, using one of the 6 classifiers evaluated based on AUC score. The configuration for each classifier is obtained by sample points from independent uniform distributions, where the respective support for each parameter is shown in Table~\ref{tab:hyperparameters_details}.

\section{Results and Discussion} \label{Sec:Results}

Our  experiments  aim  to  examine  the  following: (1) importance of each hyperparameter to each of the studied machine learning models in Section~\ref{Sec:ImportHyperParm}, (2) good defaults for each of the six studied machine learning in Section~\ref{Sec:RecommendedHyper}, (3) performance and tunability of the studied algorithms in Section~\ref{Sec:Tunability} and Section~\ref{Sec:PerformanceML}, respectively.

\subsection{Importance of Hyperparameters} \label{Sec:ImportHyperParm}

In this section, we present the experimental results for determining the important hyperparameters of the six classifiers. This analysis is based upon the performance data of six algorithms with 898507 hyperparameters, running over the 200 datasets, using 135548 CPU days to generate. All performance data we used is publicly available \footnote{https://git.io/JOPA5}. For a given algorithm $A$ and a given dataset, we use the performance data obtained for this algorithm from our knowledge base and then functional ANOVA's random forest is fit on the performance data. Next, we return the variance contribution of each hyperparameter, such that high values of variance indicate high importance. In the following, we study the distribution of these variance contributions across 200 datasets to determine that most important hyperparameters based on empirical data.

We present the results of each of the six classifiers as a set of six figures. Each figure shows the violin plots of the variance contribution of each hyperparameter across all datasets. The $x$-axis shows the hyperparameter(s) to be investigated and the $y$-axis shows the marginal contribution of the hyperparameter(s) across all datasets. A high value for the marginal contribution implies that this hyperparameter contributes for large fraction of variance, and therefore, would account for high accuracy-loss if not tuned in a proper way. 

\begin{figure} 
\includegraphics[width=0.5\textwidth, height=5cm] {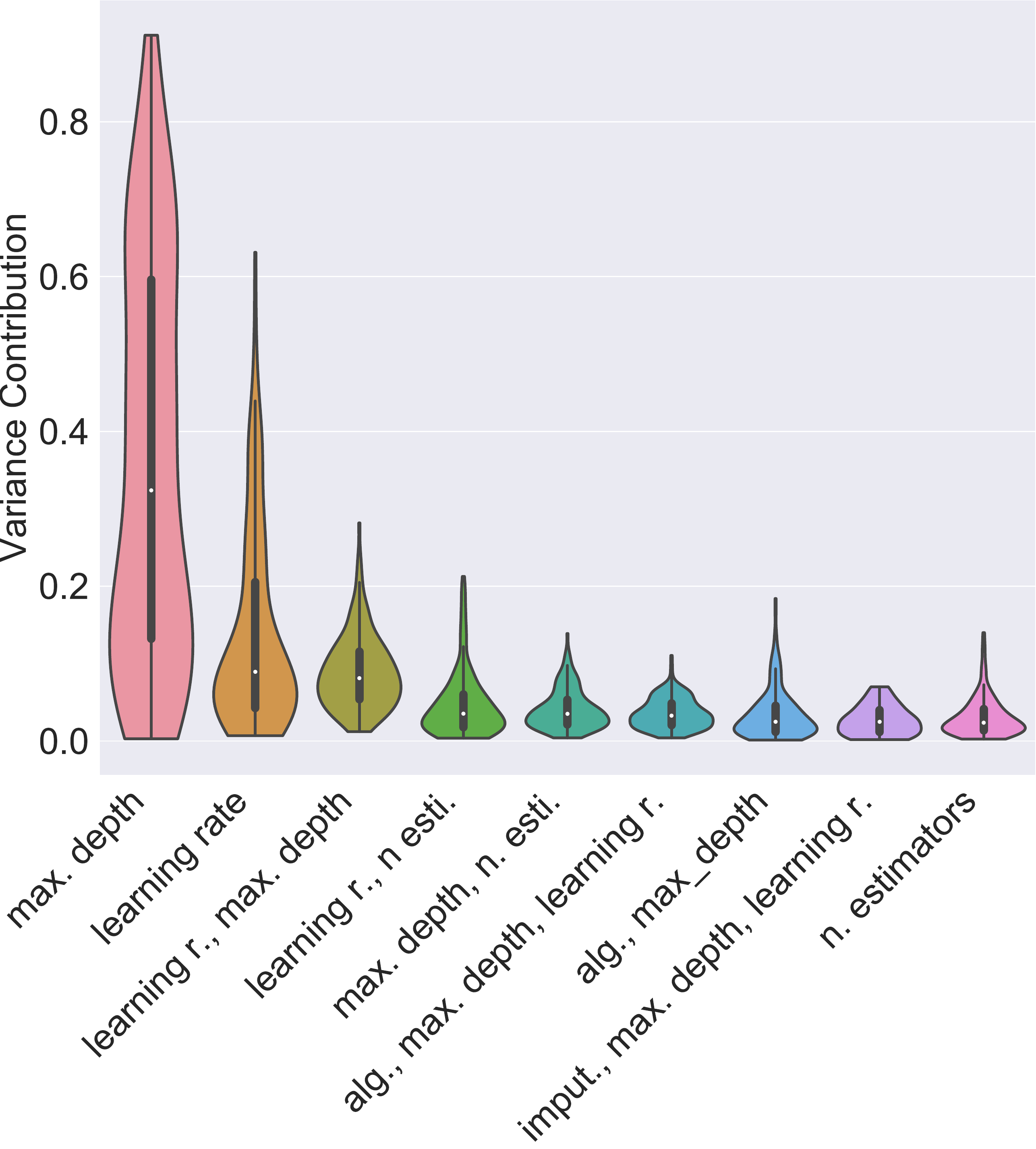}
 \caption{ Marginal contribution per 200 datasets for Adaboost classifier}
\label{Fig:HyperParamterImportanceAdaRand}
\end{figure}

\textbf{Adaboost important hyperparameters:} Figure~\ref{Fig:HyperParamterImportanceAdaRand} shows the hyperparameter importance for Adaboost. The results reveal that most of the variance could be attributed to a small set of hyperparameters. The maximal depth of the Adaboost contributes to the largest variance followed by learning rate. Both of these hyperparameters were significantly more important than the others according to the Wilcoxon signed-rank test with more  than  95\%  level  of  confidence  (p-value<0.05).

\begin{figure} 
\includegraphics[width=0.5\textwidth, height=5cm] {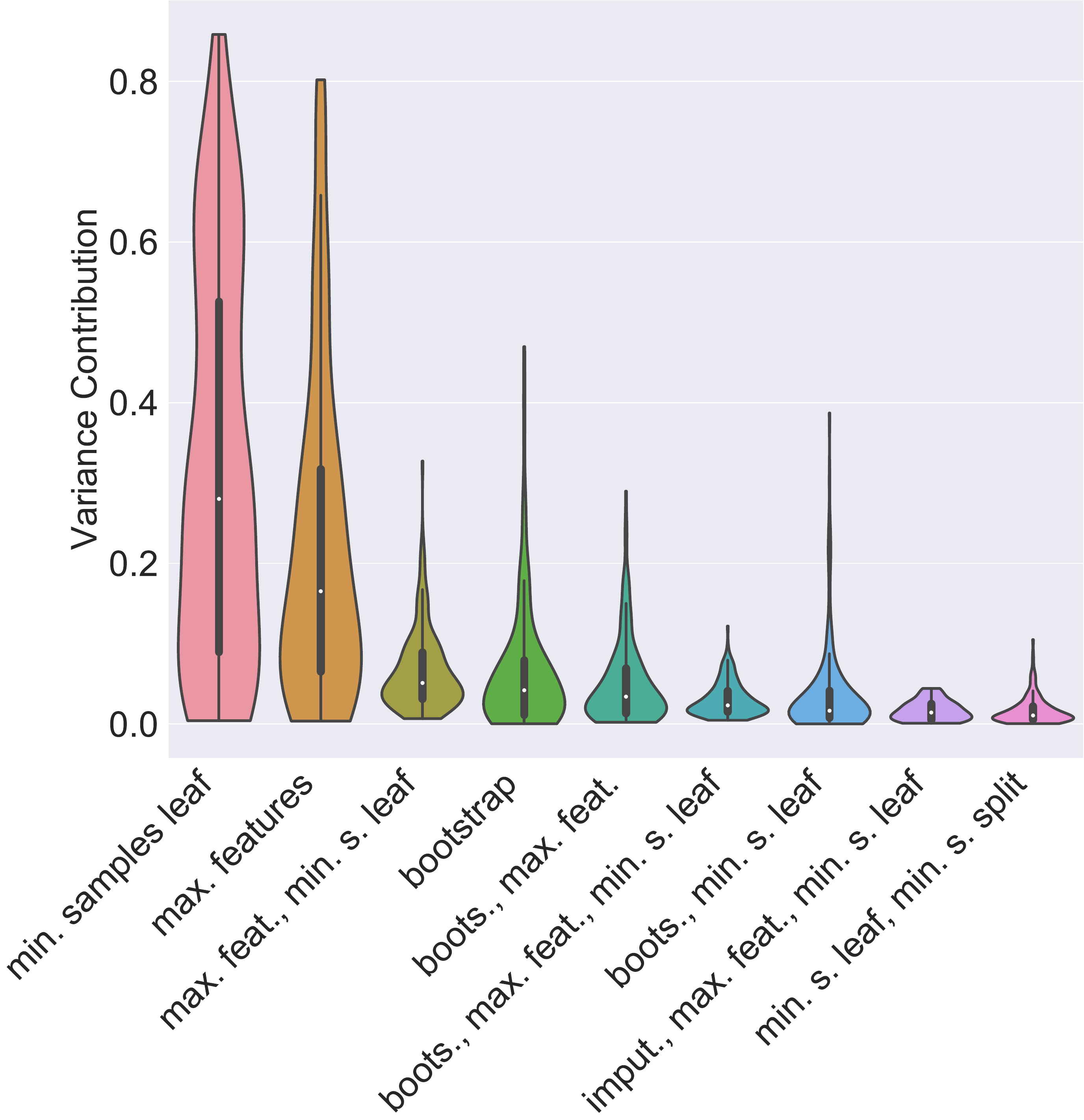}
 \caption{ Marginal contribution per 200 datasets for  random forest classifier}
 \label{Fig:HyperParamterImportanceRF}
\end{figure}

\begin{figure} 
\includegraphics[width=0.5\textwidth, height=5cm] {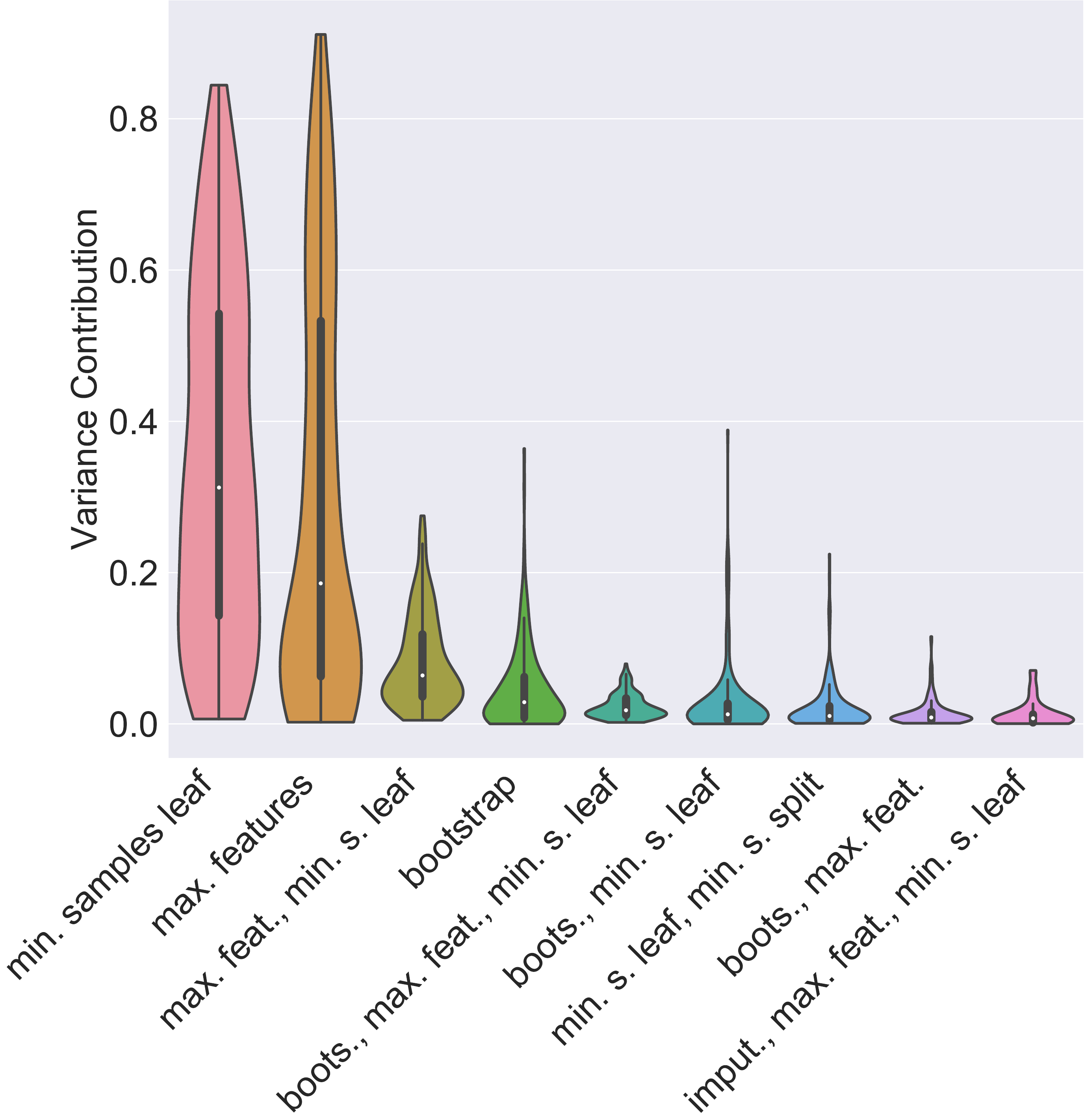}
\caption{Marginal contribution per 200 datasets for Extra Trees classifier}
\label{Fig:HyperParamterImportanceET}
\end{figure}

\textbf{Random forest and Extra Trees important hyperparameters:} Figure~\ref{Fig:HyperParamterImportanceRF} and Figure~\ref{Fig:HyperParamterImportanceET} show that the most contributed hyperparameters to the variance for random forest and Extra Trees algorithms, respectively. The results show that the most contributed hyperparameters to the variance are the minimum samples per leaf and maximal number of features for determining the split. The results obtained from functional ANOVA aligned with the findings that concluded that ensembles perform well when the individual models outperform the random guessing and when the errors obtained by the individual models are uncorrelated ~\cite{zhou2002neural,breiman2001random}. The interaction between the minimum samples per leaf and maximal number of features for determining the split is significantly more important than the rest of examined interactions between hyperparameters. Based on the Wilcoxon signed-rank test, both hyperparameters were significantly more important than the rest of hyperparameters with more  than  95\%  level  of  confidence  (p-value<0.05).

\begin{figure} 
\includegraphics[width=0.5\textwidth, height=5cm] {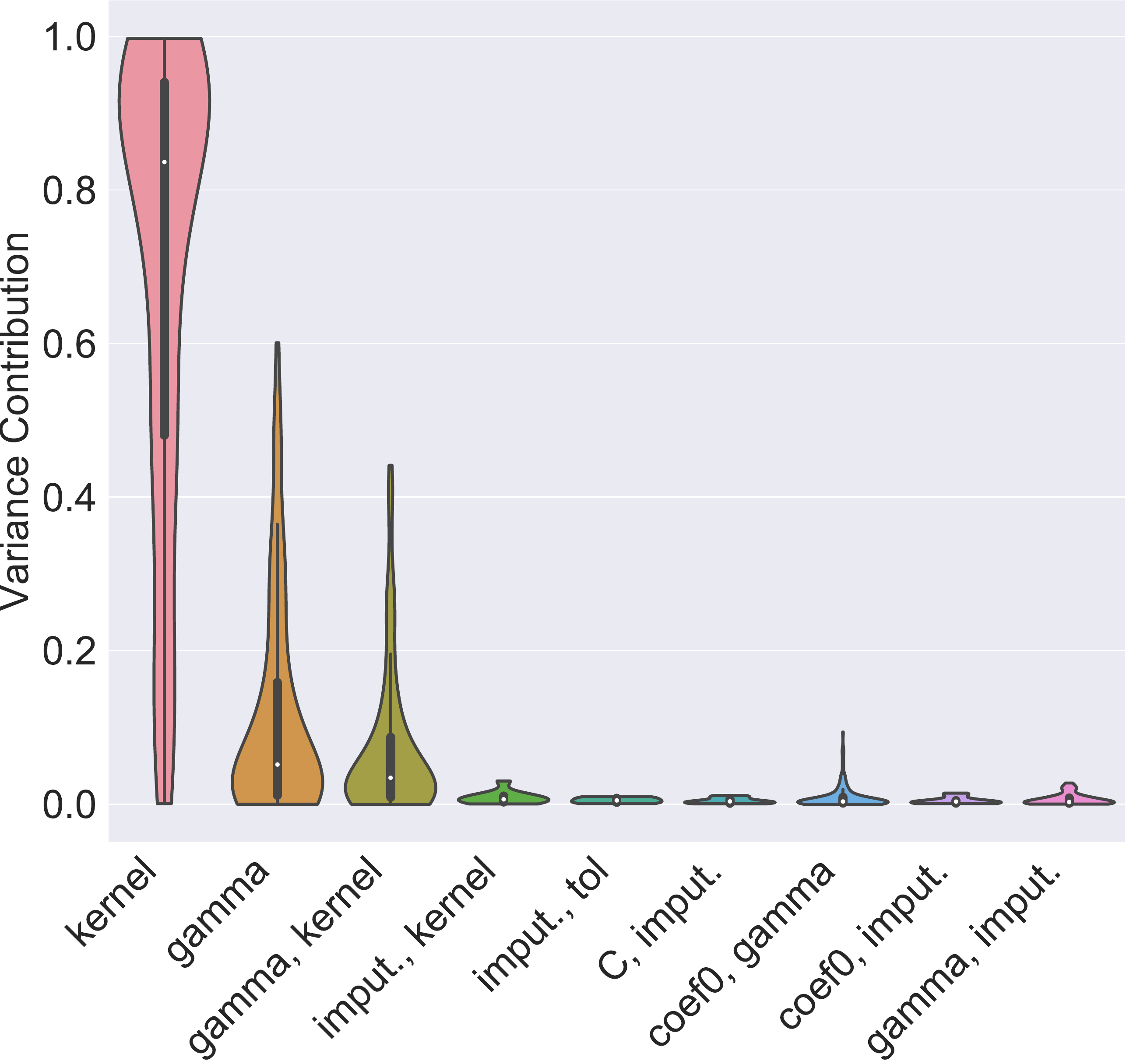}
\caption{  Marginal contribution per 200 datasets for SVMs classifier}
\label{Fig:HyperParamterImportanceSVM}
\end{figure}

 \textbf{SVMs important hyperparameters:} Figure~\ref{Fig:HyperParamterImportanceSVM} shows the analysis results of the SVMs. The results reveal that the most variance could be attributed clearly to the kernel type and gamma hyperparameters. Both of these hyperparameters are significantly more important than any other hyperparameters based on the Wilcoxon signed-rank test. Figure~\ref{Fig:HyperParamterImportanceSVM} shows that the interaction between the kernel function and the gamma is significantly more important than the rest of examined interactions between hyperparameters with more  than  95\%  level  of  confidence  (p-value<0.05).

\begin{figure} 
\includegraphics[width=0.5\textwidth, height=5cm] {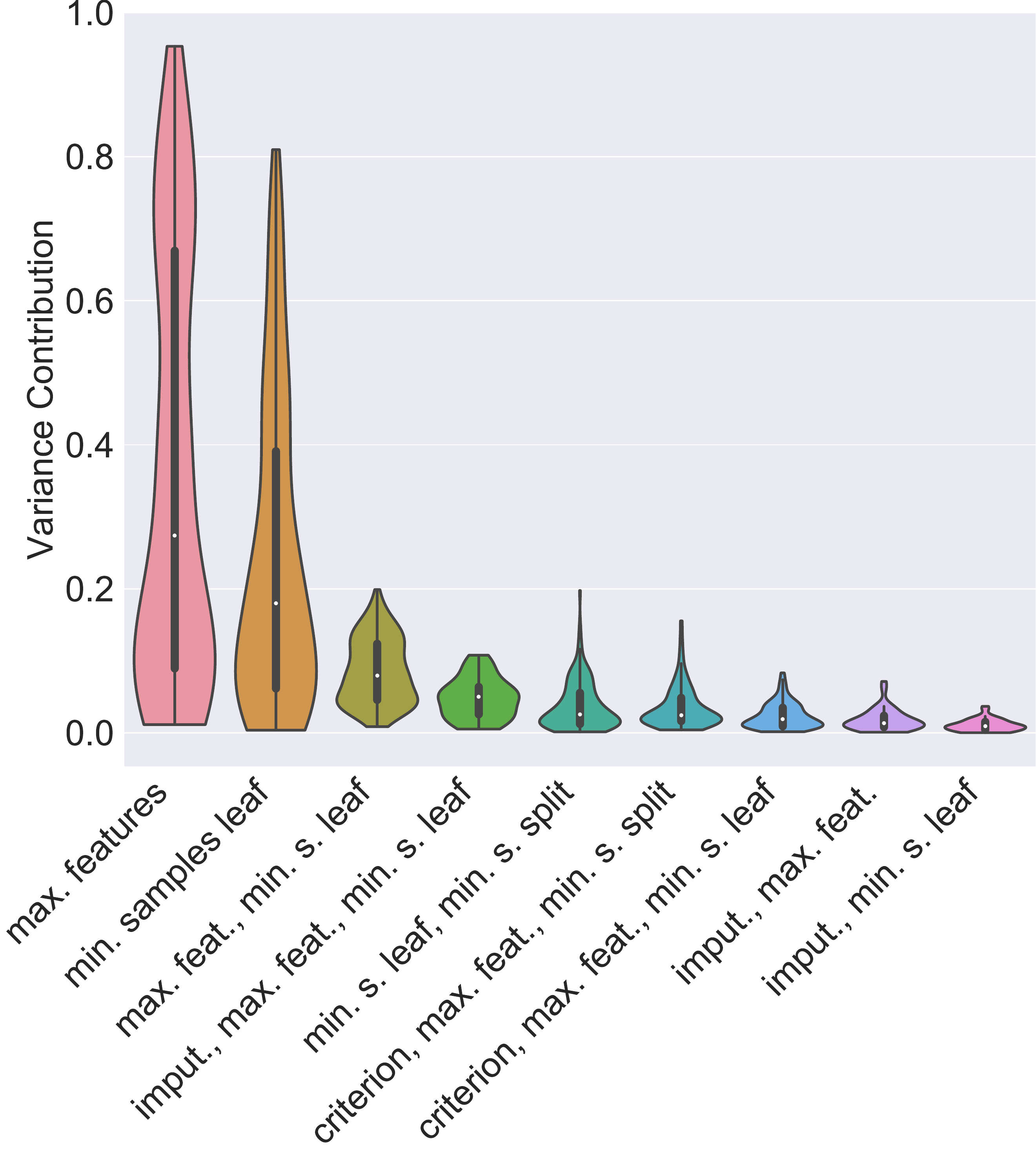}
\caption{Marginal contribution per 200 datasets for decision tree classifier}
\label{Fig:HyperParamterImportanceDT}
\end{figure}

 \textbf{Decision tree important hyperparameters:} Figure~\ref{Fig:HyperParamterImportanceDT} shows the analysis results of the decision tree algorithm. The results reveal that the hyperparameters that contributed the most to the variance are the maximal number of features for determining the split and the minimum samples per leaf. Both of these hyperparameters are statistically more significant than any other hyperparameters based on the Wilcoxon signed-rank test with more  than  95\%  level  of  confidence  (p-value<0.05). Figure~\ref{Fig:HyperParamterImportanceDT} shows that the interaction between the maximal number of features for determining the split and the minimum samples per leaf are statistically more important than the rest of examined interactions between hyperparameters, followed by the interaction between the minimum number of samples required to be at a leaf node and the number of features to consider when looking for the best split as shown in Figure~\ref{Fig:HyperParamterImportanceDT}.
 
\begin{figure} 
\includegraphics[width=0.5\textwidth, height=5cm] {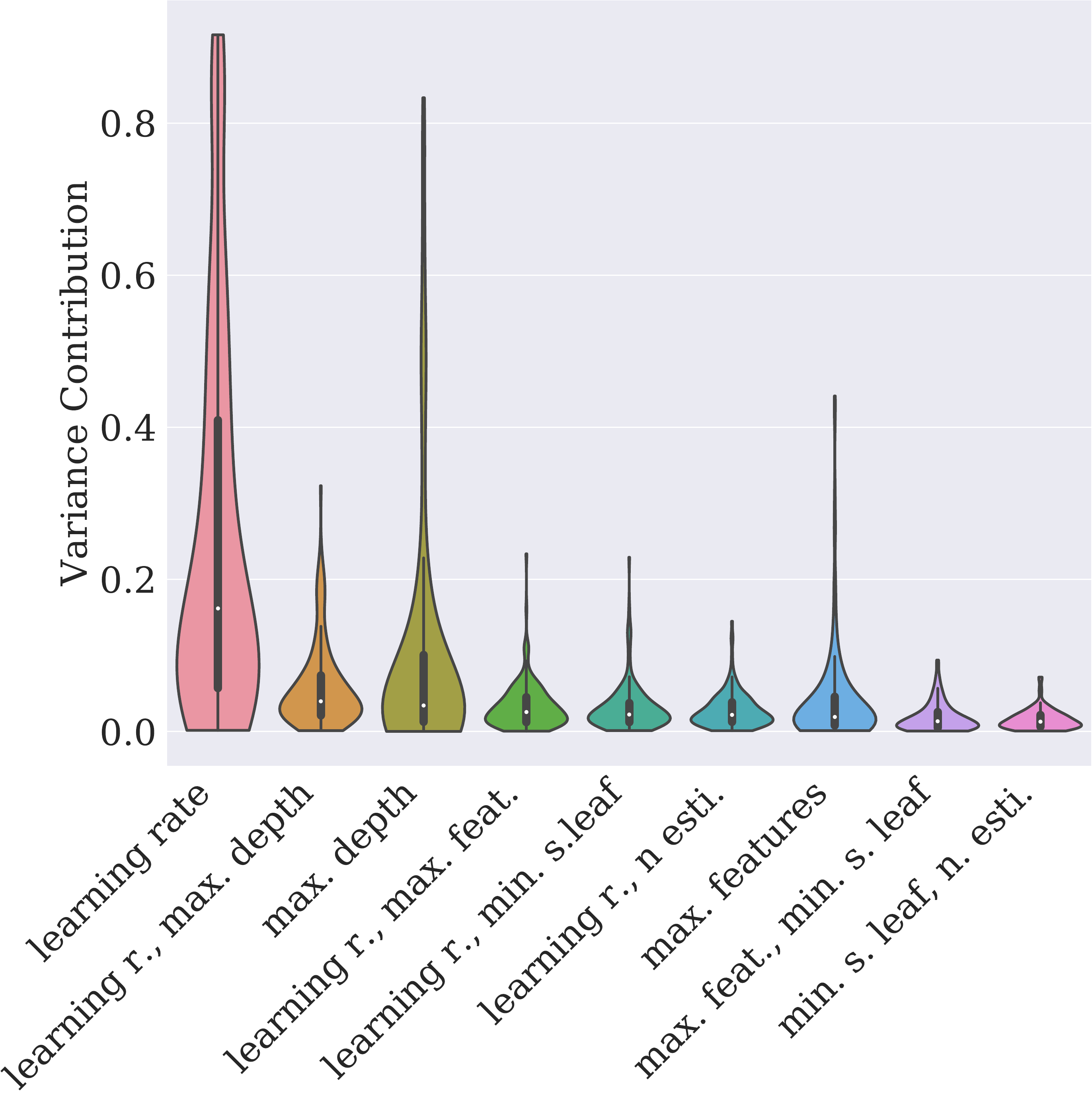}
\caption{  Marginal contribution per 200 datasets for  gradient boosting classifier}
\label{Fig:HyperParamterImportanceGB}
\end{figure}

 \textbf{Gradient Boosting important hyperparameters:} Figure~\ref{Fig:HyperParamterImportanceGB} shows the analysis results of the gradient boosting algorithm. The results show that the hyperparameter contributing the most to the variance is the learning rate. The interaction between the learning rate and the maximum depth is significantly more important than the maximum depth hyperparameter, as shown in Figure~\ref{Fig:HyperParamterImportanceGB}. Based on the Wilcoxon signed-rank test, learning rate and the interaction between these two hyperparameters were significantly more important than the rest of hyperparameters with more  than  95\%  level  of  confidence  (p-value<0.05). In addition, the hyperparameters that specify the minimum number of data points for making a leaf and the subset of features when trying to find the best split are statistically more important than others with more than  95\%  level  of  confidence  (p-value<0.05).

We compare the importance of hyperparameters reported in this study for each of Adaboost, and Random forest to the results obtained in~\cite{van2018hyperparameter,van2017empirical}. The results show that for these two algorithms, the top two hyperparameters obtained in this study comes inline with the reported results in~\cite{van2018hyperparameter,van2017empirical}.

 \begin{figure*}
\begin{multicols}{4}
    \includegraphics[width=0.25\textwidth]{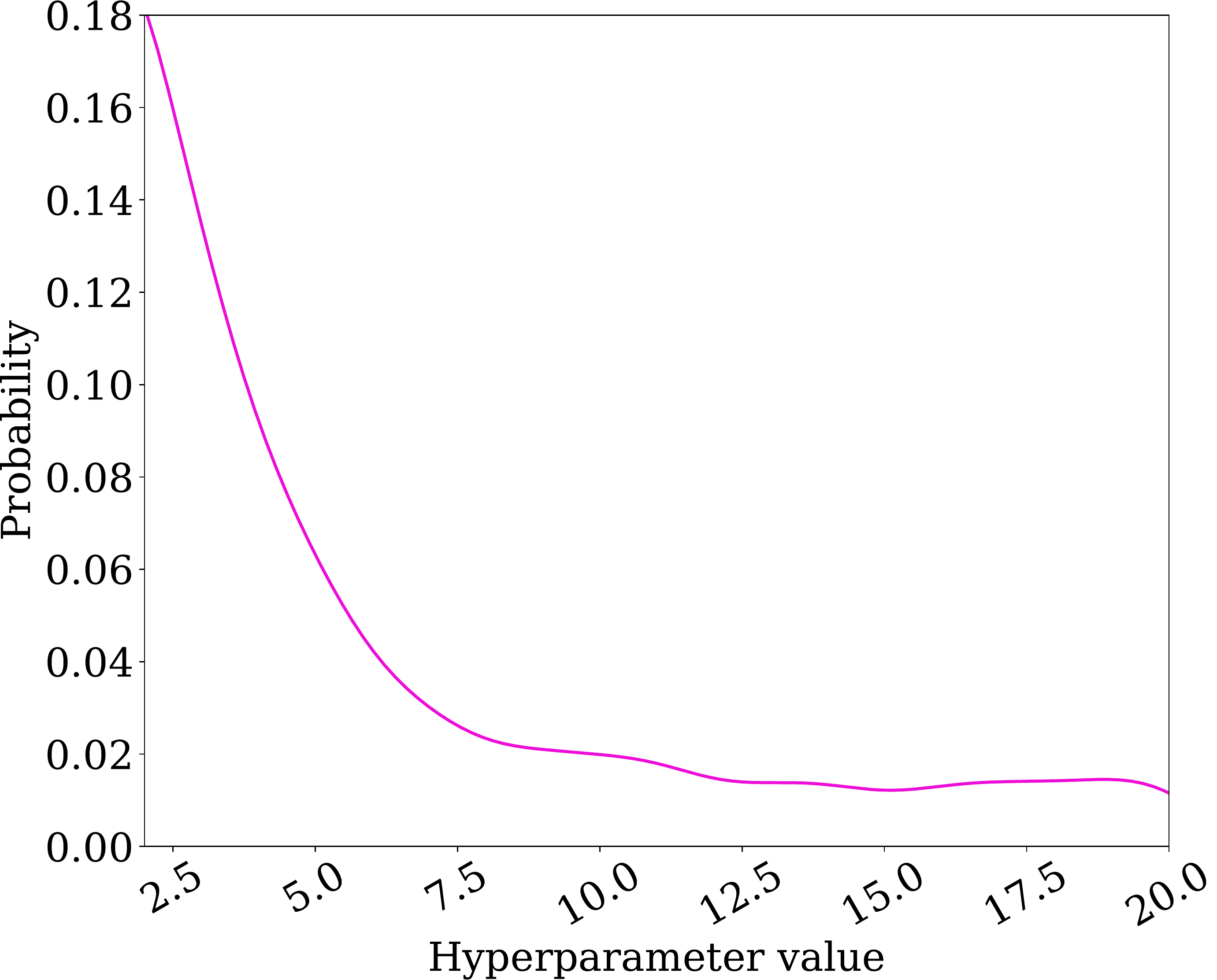}\par\subcaption{RF: min. samples leaf}
    \label{fig:best_value1}
        \includegraphics[width=0.25\textwidth]{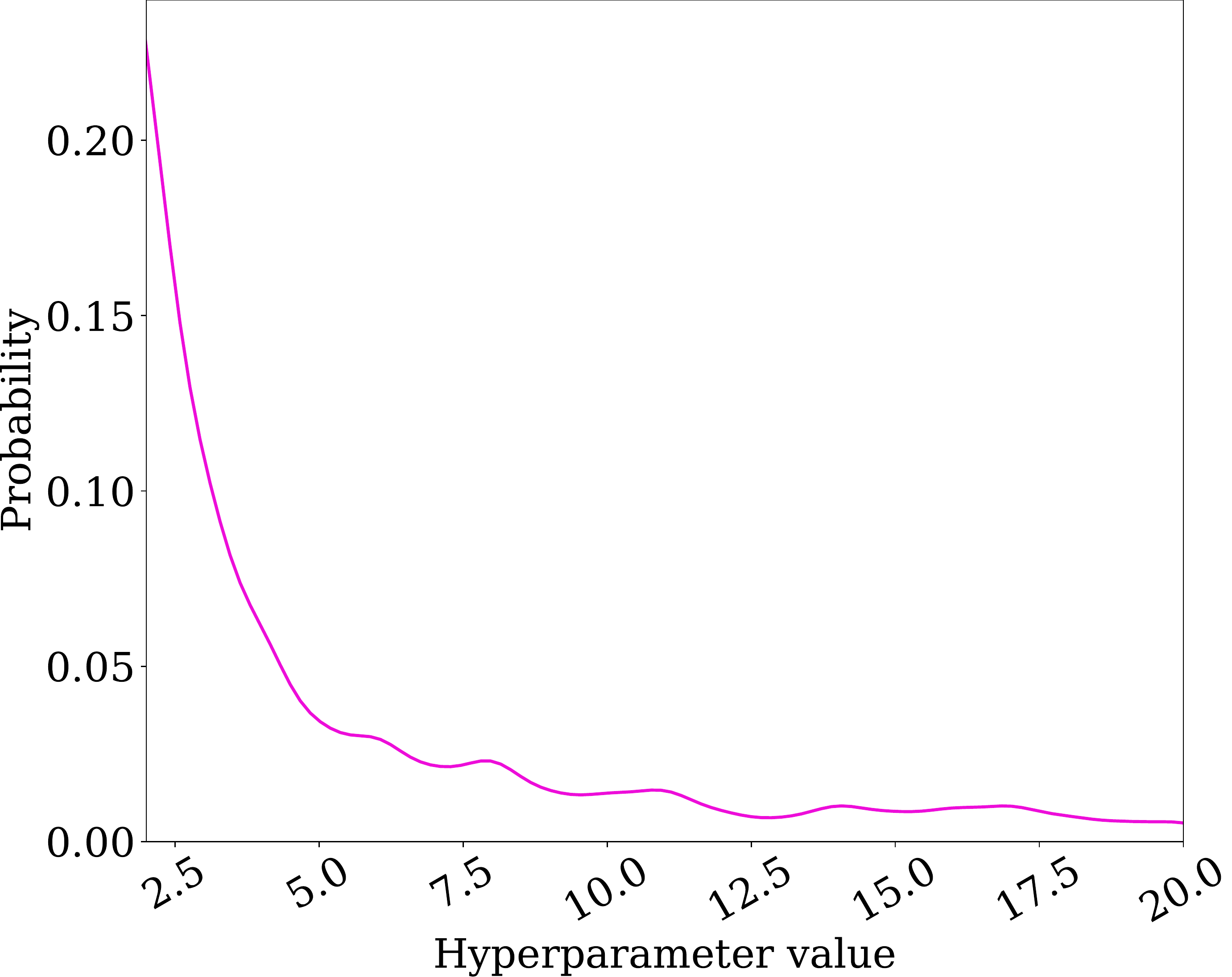}\par\subcaption{ET: min. samples leaf} 
  
    \label{fig:best_value2}
    \includegraphics[width=0.25\textwidth]{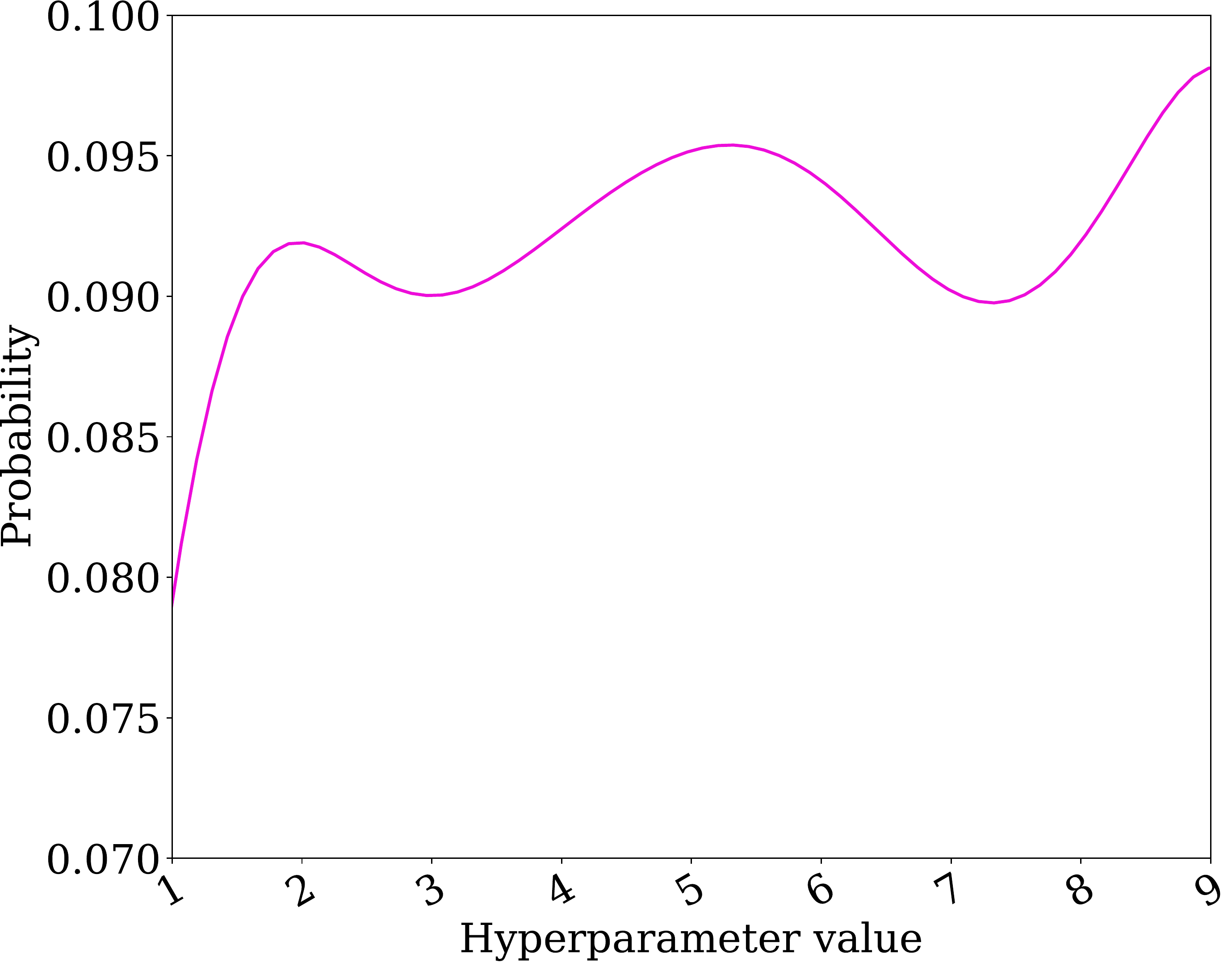}\par\subcaption{Adaboost: max. depth} 
   
    \label{fig:best_value3}
        \includegraphics[width=0.25\textwidth]{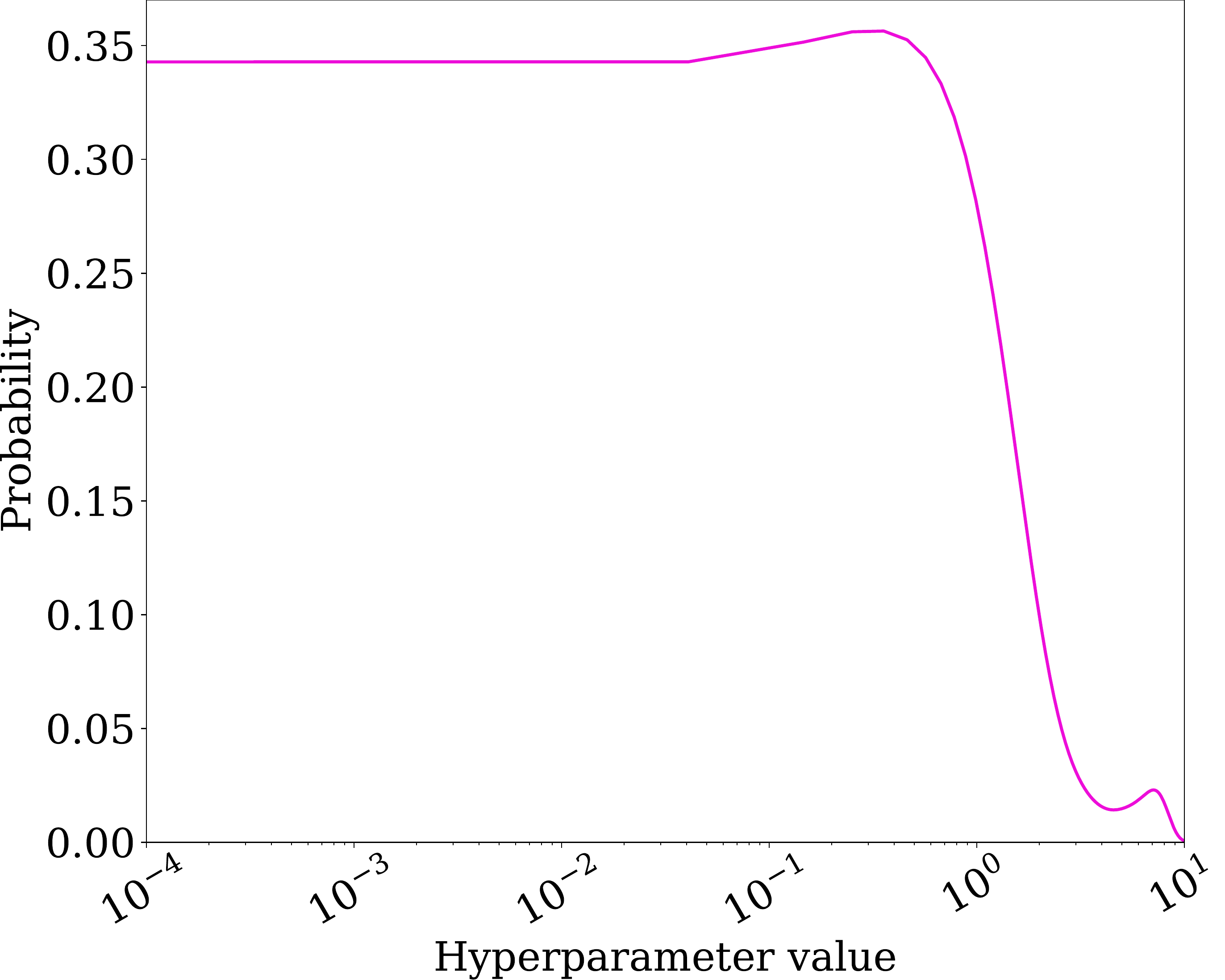}\par\subcaption{SVM(rbf kernel):gamma}
 
    \label{fig:best_value4}
\end{multicols}
\begin{multicols}{3}
    \includegraphics[width=0.25\textwidth]{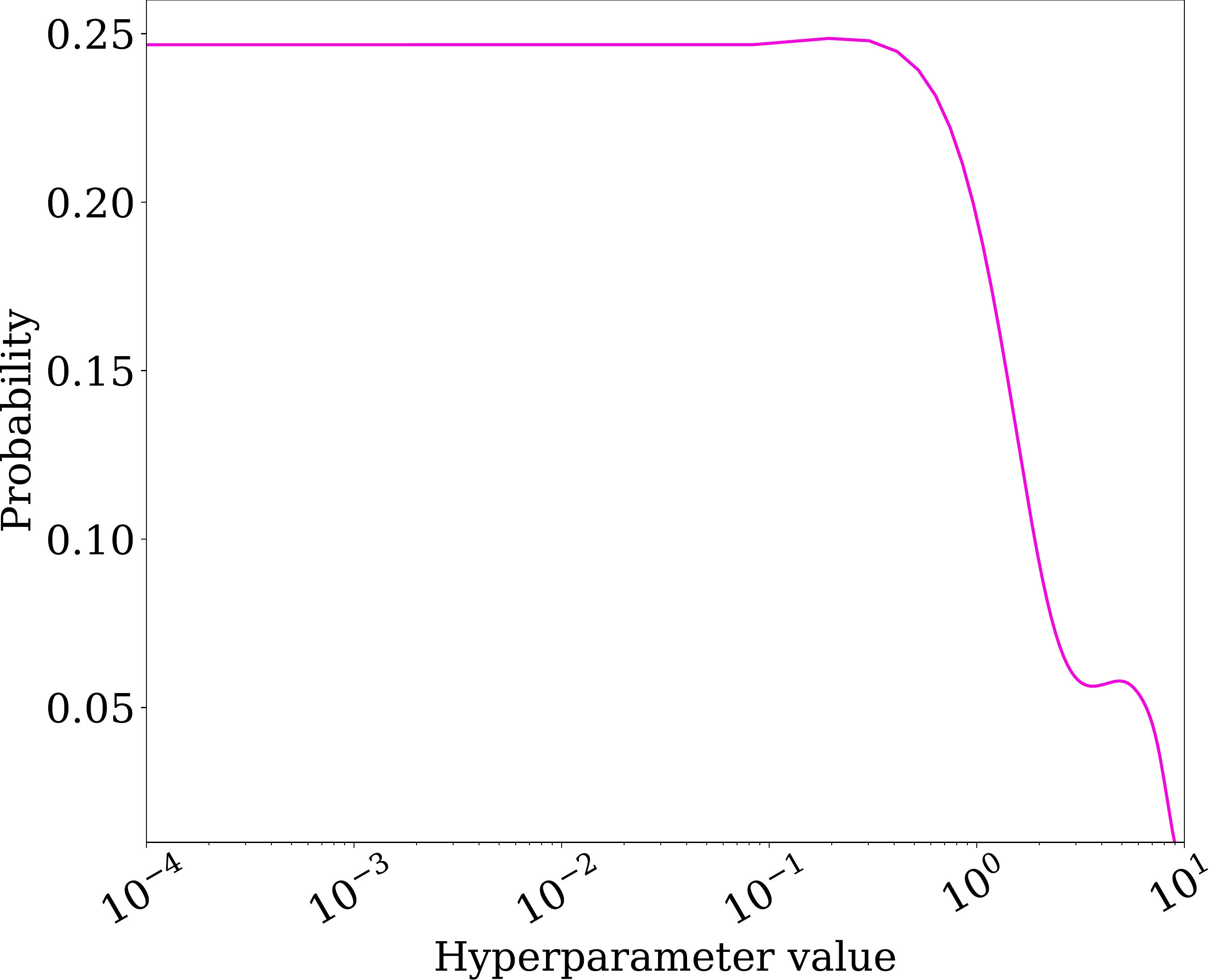}\par\subcaption{SVM(sigmoid kernel):gamma} 
    \label{fig:best_value5}
          \includegraphics[width=0.25\textwidth]{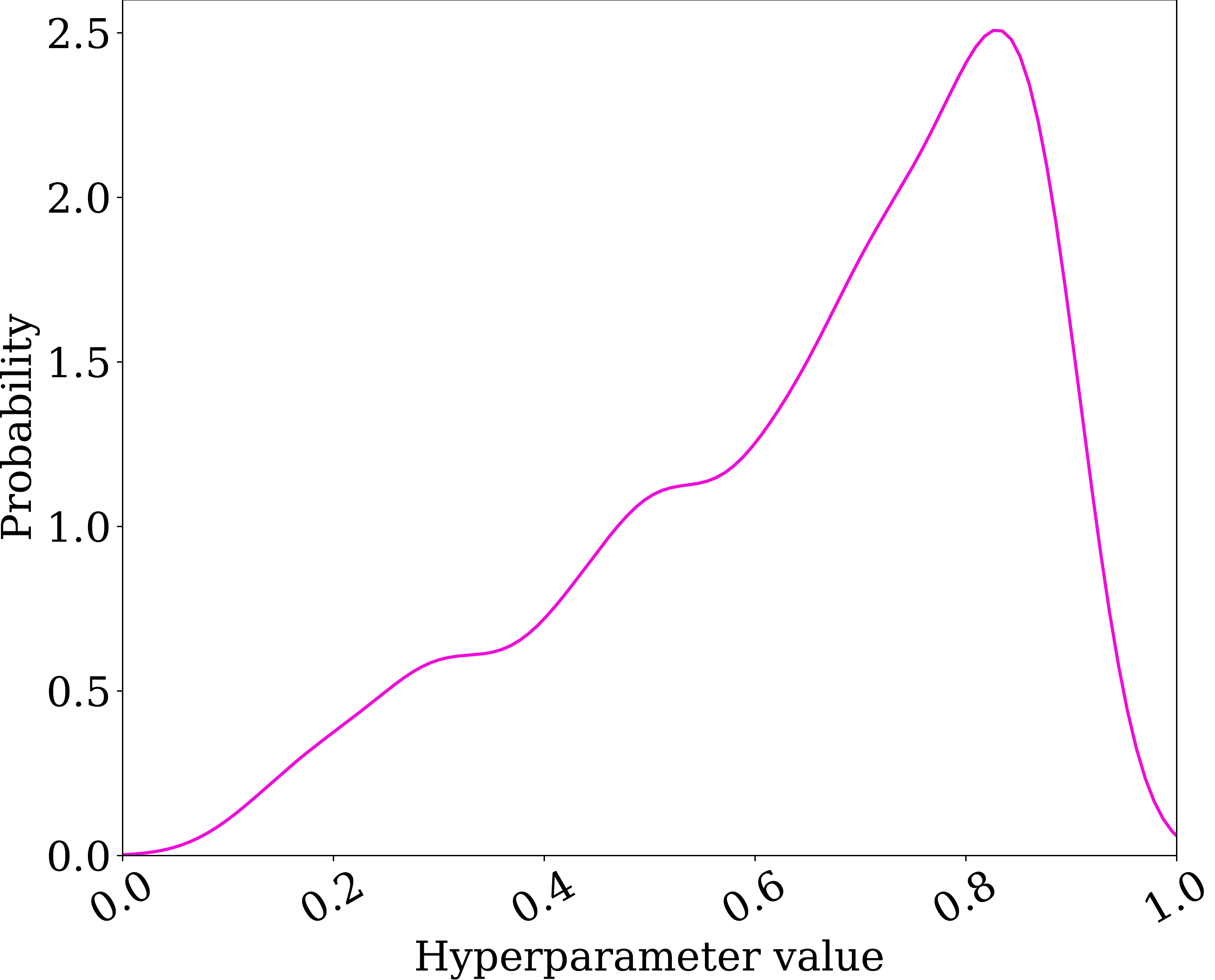}\par\subcaption{Decision tree: max.feature}
    \label{fig:best_value6}
     \includegraphics[width=0.25\textwidth]{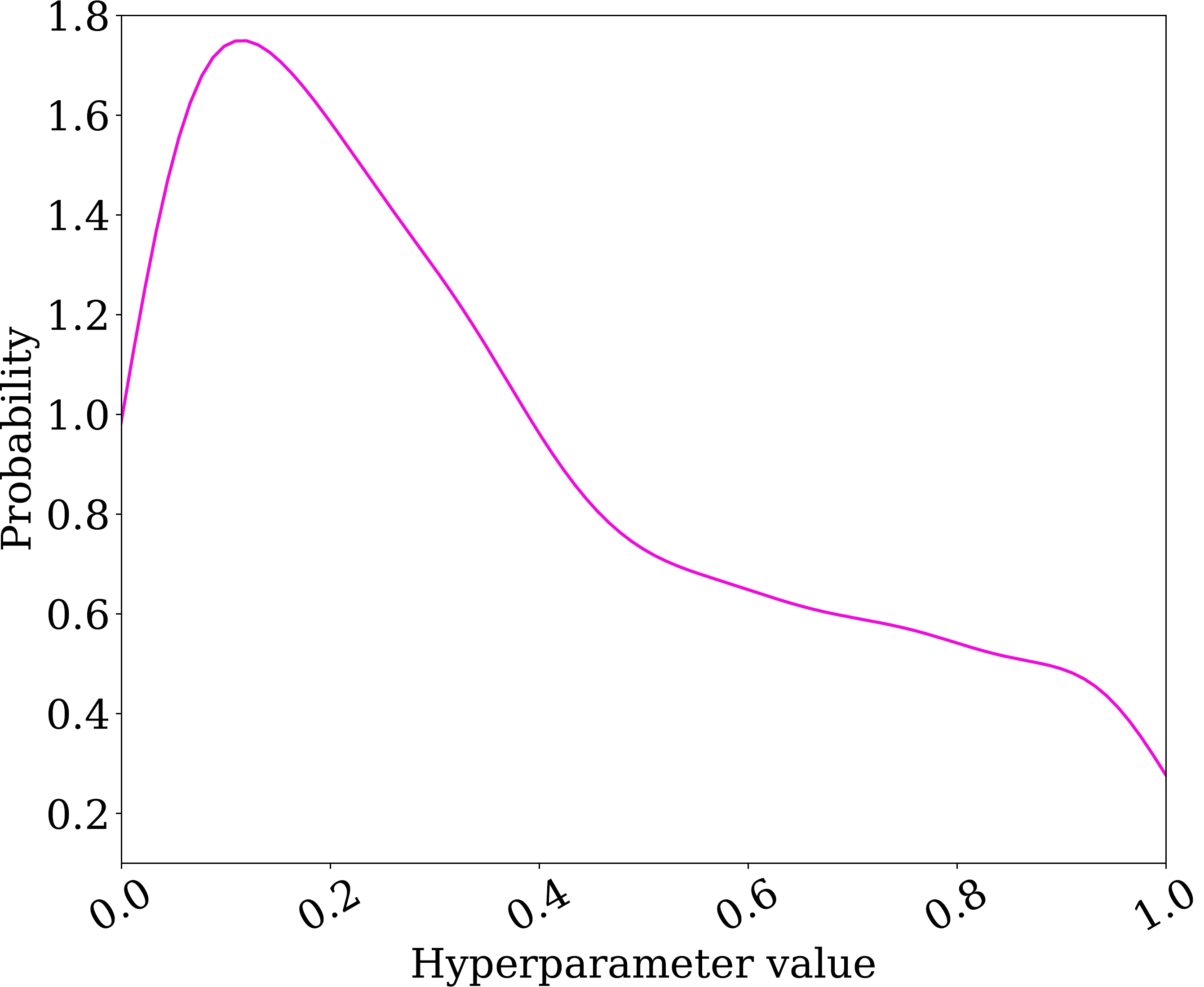}\par\subcaption{GB: learning rate} 

\label{fig:best_value7}
    
\end{multicols}
\caption{Obtained Priors for the hyperparameter found to be most important for each classifiers over 200 datasets. Th x-axis represents the value of the hyperparameter, the y-axis represents the probability that this value will be sampled }\label{fig:best_value}
\end{figure*}
 
\subsection{Recommended Hyperparameter values}\label{Sec:RecommendedHyper}
 In this section, we focus on recommending values for hyperparameters that tended to yield good performance. Figure \ref{fig:best_value} shows the kernel density estimators for the most important hyperparameters for each of the six classifiers included in this study, where the x-axis represents the hyperparameter value and the y-axis represents the probability that this value will be sampled. Figure \ref{fig:best_value1} and Figure \ref{fig:best_value2} show that the hyperparameter represents the minimum number of samples required to be at a leaf node for the random forest and Extra Trees models should be set to a small value, and hence, the default value of 1 for this hyperparameter is quite good, which  is consistent with the findings in ~\cite{geurts2006extremely}. For Adaboost classifier, the maximum depth should be set to a large value, as shown in Figure \ref{fig:best_value3}. For both types of SVMs, the gamma hyperparameter should be set to be less than 1 to achieve good performance, as shown in Figure \ref{fig:best_value4} and Figure \ref{fig:best_value5}. For decision tree classifier, the number of features to consider when looking for the best split should be set to a large value less than 0.8, as shown in Figure \ref{fig:best_value6}. The learning rate hyperparameter for gradient boosting algorithm has a good default and should be set to a small value to achieve a good performance as shown in Figure \ref{fig:best_value7}.

\begin{figure}[t]
\includegraphics[width=0.4\textwidth, height=5cm] {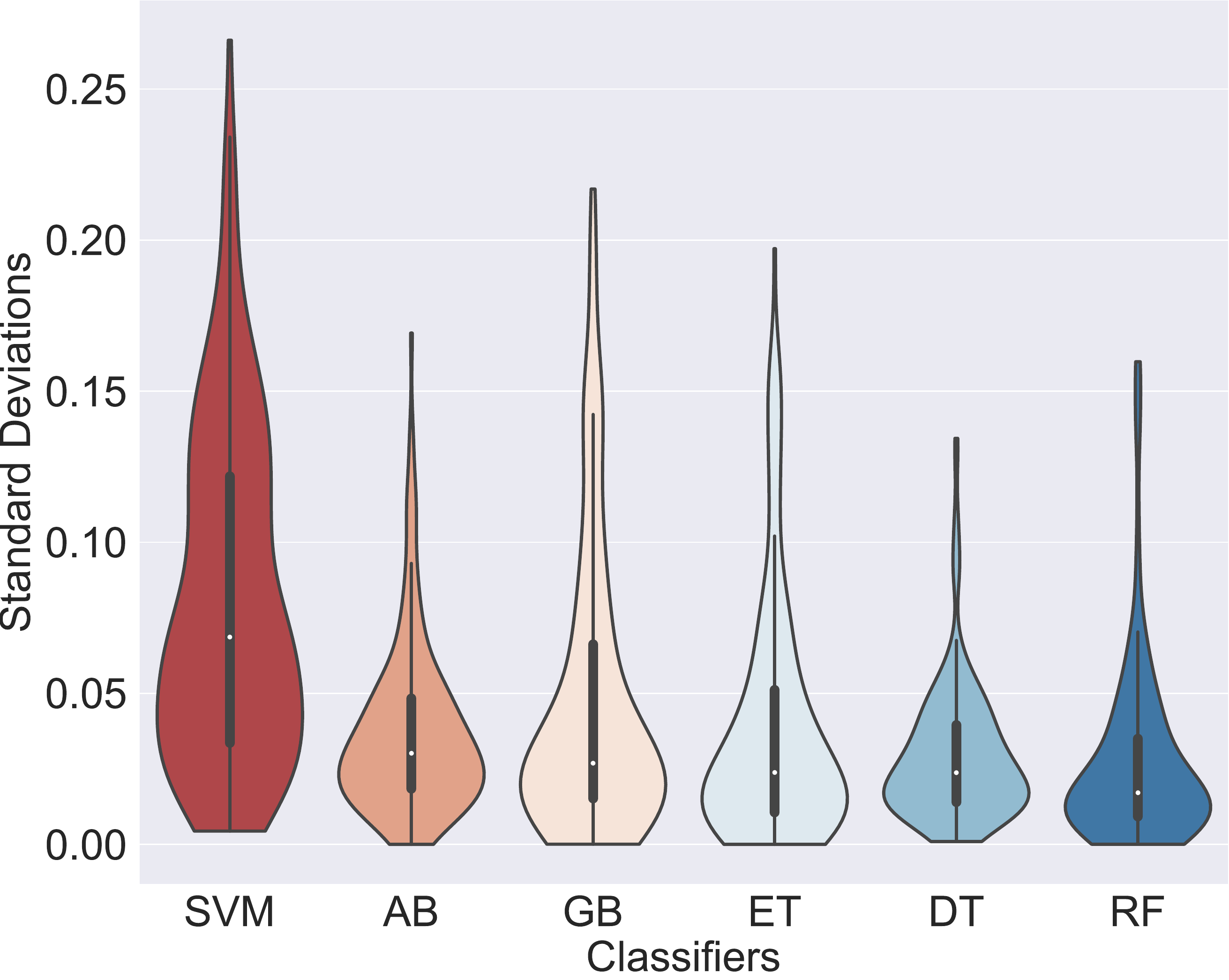}
\caption{Tunability results for each algorithm evaluated on 500 configurations for each of the 200 datasets included in this study. }
\label{fig:standard_deviation}
\end{figure}

\subsection{Measuring overall tunability of a Machine Learning algorithm}\label{Sec:Tunability}

A general measure of the tunability of a machine learning algorithm per dataset can be computed based on the difference between the performance of an overall reference configuration and the performance of the best possible configuration on that dataset. For each algorithm, this gives rise to an empirical distribution of performance differences over datasets, which might be directly summarized to an aggregated tunability measured by using mean, or standard deviation. In this work, we use the standard deviation to assess the tunability of each classifier. Figure \ref{fig:standard_deviation} shows the tunability results for each algorithm evaluated on 500 configurations for each of the 200 dataset included in this study. Clearly, some algorithms such as SVMs, gradient boosting and Extra Trees are much more tunable than the others, while decision tree and random forest is the least tunable one. This results are consistent with the general knowledge in machine learning and with the research findings in~\cite{probst2019tunability}.

\begin{figure}               
\includegraphics[width=0.3\textwidth, height=5cm] {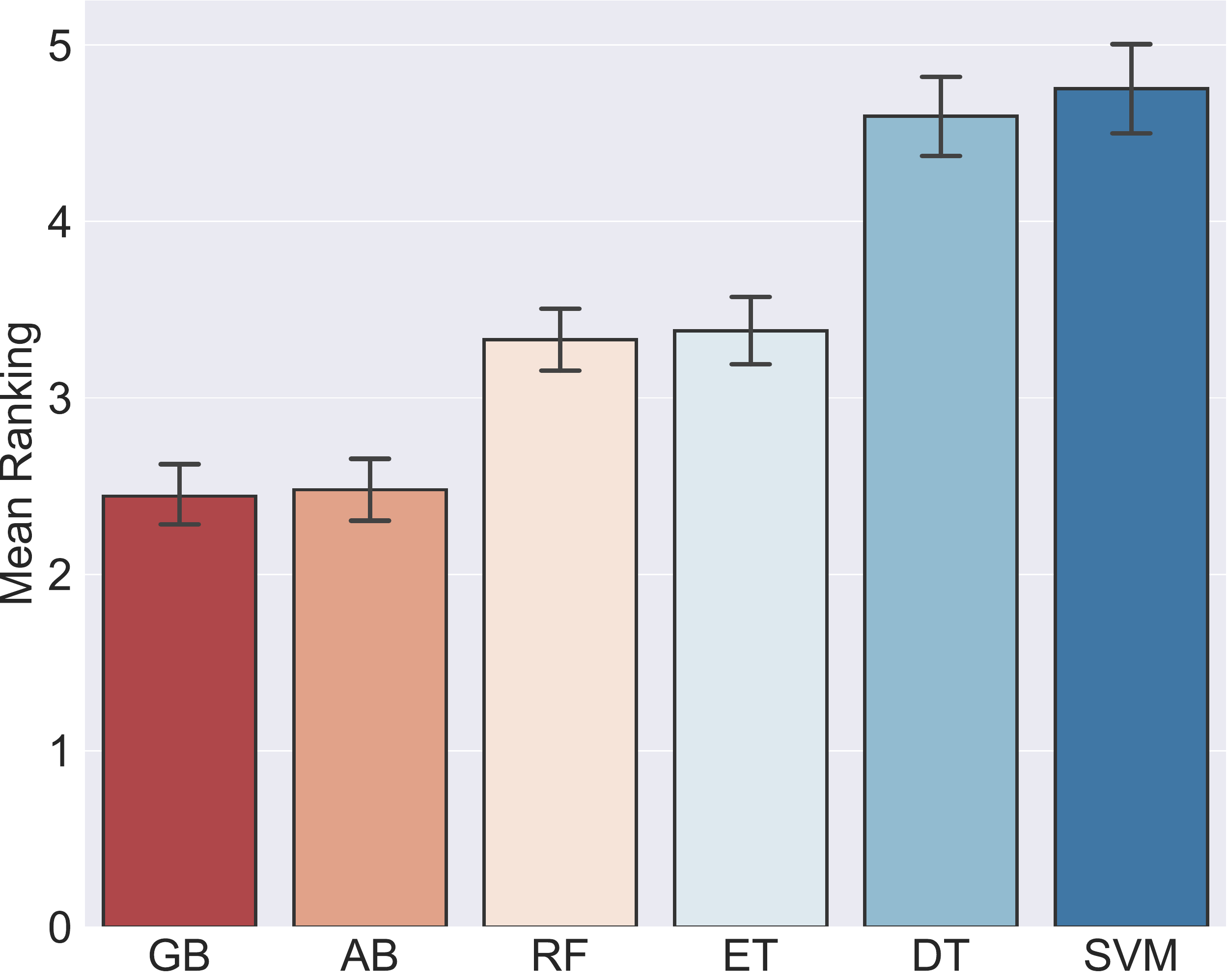}
\caption {Mean ranking of six Machine Learning algorithms over 194 datasets. Error bars indicate the 95\% confidence interval.}
\label{Fig:MeanRank}
\end{figure}

\begin{figure}               
\includegraphics[width=0.3\textwidth, height=5cm] {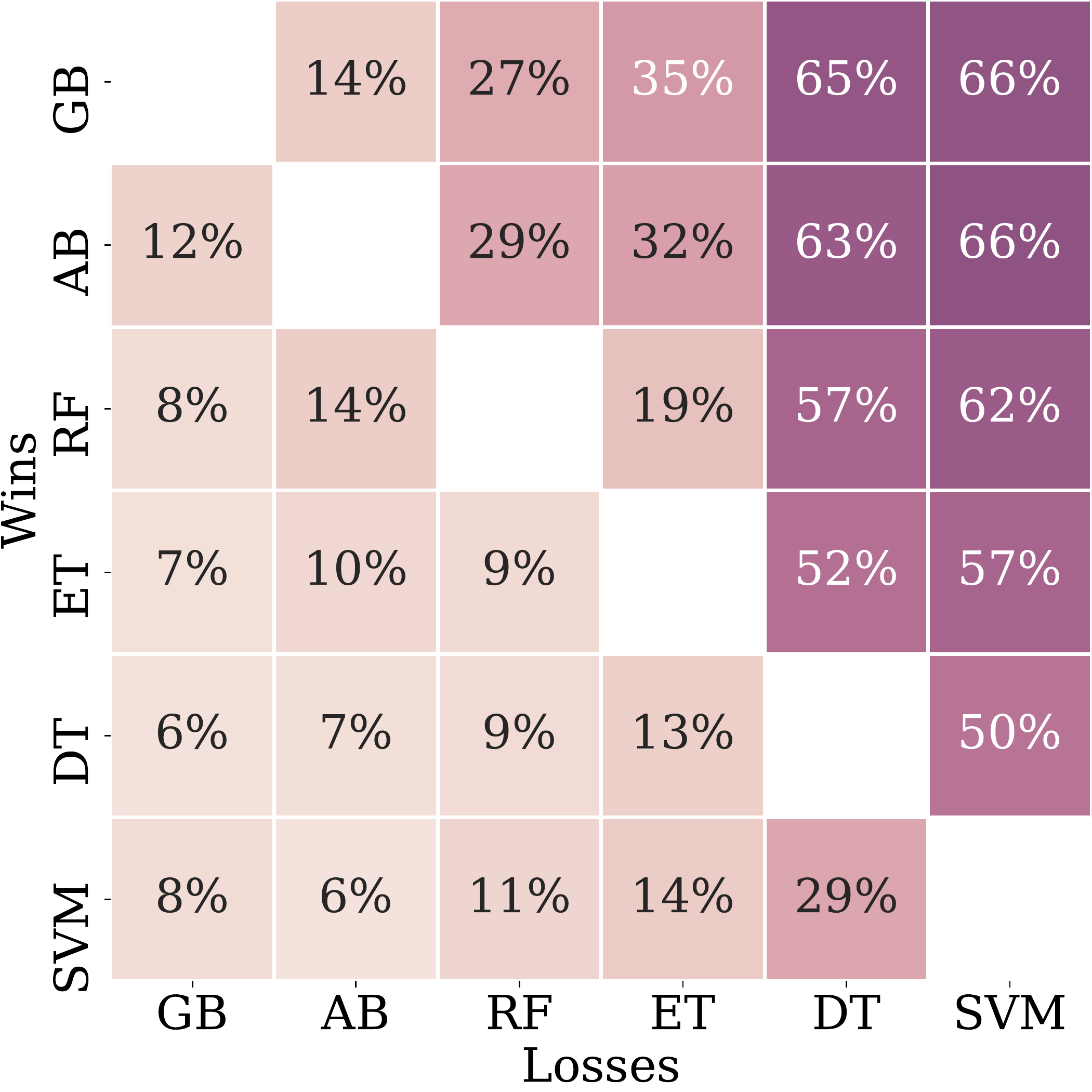}
\caption {The percentage of a given algorithm out of 194 datasets outperforms another algorithm in terms of the best AUC on a problem. Algorithms are ordered from top to bottom on the basis of their overall performance on all datasets. Two algorithms are considered to have same performance on a problem when they have achieved accuracy within 1\% of each other.}
\label{Fig:PercentageWin}
\end{figure}

\subsection{Measuring overall performance of a Machine Learning algorithm }\label{Sec:PerformanceML}
 
In this section, we compare the performance of each algorithm across all datasets. Figure~\ref{Fig:MeanRank} compares the performance of the six algorithms included in this study, we plot the mean rankings of the algorithms across all datasets. The rankings show the strength of ensemble-based tree algorithms in generating highly performing models. Gradient boosting is the first ranked algorithm, followed by Adaboost, while SVMs has got the worst ranking. In order to assess the statistical significance of the observed differences in algorithm performance across all problems, we use the Wilcoxon signed-rank test. The results show that the difference in performance between all classifiers is statistically significant  with more than 95\% level of confidence (p-value<0.05) except that between Gradient boosting and Adaboost, and between Extra Trees and Random forest. Given the huge amount of obtained performance data, we try to recommend the top-ranked algorithms across the 200 datasets. However, it is worth mentioning that the top-ranked algorithms may not outperform others for some problems. Furthermore, when simpler algorithms perform on par with a more complex one, it is often preferable to choose the simpler of the two. We investigate the pair-wise performance comparison between classifiers by calculating the percentage of datasets for which one algorithm outperforms another, shown in Figure~\ref{Fig:PercentageWin}. It is clear from the results that there is no single machine model that achieves the highest performance across all datasets. For example, there are 16 datasets for which SVMs performs as well as or better than gradient boosting, despite being the overall worst- and best-ranked algorithms, respectively.

\section{Conclusions and Future work}\label{SEC:Conclusion}

In this work, we study the hyperparameters of different machine learning techniques and the impact of tuning them either jointly, tuning individual parameters or combinations, all based on the general concept of surrogate empirical performance models and ANOVA framework. More specifically, we identified the main important hyperparameters for six machine learning algorithms including Adaboost, random forest, Extra Trees, SVMs, decision tree, and gradient boosting. In addition, we quantify the tunability of such algorithms, which have never been provided before in such a principled manner based on a large number of datasets and based on a wide range of problems, including binary and multiclass classification problems. Our results conform with common knowledge and results from literature. We also compared the top important hyperparameters for random forest and Adaboost to those obtained from van Rijn and Hutter~\cite{van2017empirical}.

One future direction is to study the hyperparameters for different machine learning problems, including regression and clustering algorithms. Finally, we aim to employ recent advances in meta-learning to identify similar datasets and base the priors only on these to yield dataset-specific priors for hyperparameter optimization.


\section*{Acknowledgement}
The research specified under the object of the agreement was carried out with the support of the European Regional Development Fund and the programme Mobilitas Pluss (project number: 2014-2020.4.01.16-0024).